\documentclass{article} 
\usepackage{iclr2026_conference,times}

\usepackage{amsmath,amsfonts,bm}

\def\eqref#1{equation~\ref{#1}}

\def\1{\bm{1}}

\DeclareMathAlphabet{\mathsfit}{\encodingdefault}{\sfdefault}{m}{sl}
\SetMathAlphabet{\mathsfit}{bold}{\encodingdefault}{\sfdefault}{bx}{n}

\usepackage{hyperref}
\usepackage{url}
\usepackage{graphicx}
\usepackage{booktabs}
\usepackage{algorithm}
\usepackage{algpseudocode}
\algrenewcommand\algorithmiccomment[1]{\text{// #1}}
\usepackage{amsmath}

\usepackage{hyperref}
\usepackage{xcolor}

\title{Learning Illumination Control \\ in Diffusion Models}

\author{Nishit Anand, Manan Suri, Christopher Metzler$^*$, Dinesh Manocha$^*$, Ramani Duraiswami$^*$ \\
University of Maryland College Park \\
\texttt{Corresponding Author: nishit@umd.edu} \\
\texttt{\small{Project Website: \textcolor{magenta}{\href{https://nishitanand.github.io/relighting-diffusion-website}{nishitanand.github.io/relighting-diffusion-website}}}} \\
{\small $^*$Equal Advising}}

\iclrfinalcopy
\begin{document}

\maketitle

\begin{abstract}
Controlling illumination in images is essential for photography and visual content creation. While closed-source models have demonstrated impressive illumination control, open-source alternatives either require heavy control inputs like depth maps or do not release their data and code. We present a fully open-source and reproducible pipeline for learning illumination control in diffusion models. Our approach builds a data engine that transforms well-lit images into supervised training triplets consisting of a poorly-illuminated input image, a natural language lighting instruction, and a well-illuminated output image. We finetune a diffusion model on this data and demonstrate significant improvements over baseline SD 1.5, SDXL, and FLUX.1-dev models in perceptual similarity, structural similarity, and identity preservation. Our work provides a reproducible solution built entirely with open-source tools and publicly available data. We release all our code, data, and model weights publicly.
\end{abstract}

\section{Introduction}

Lighting is fundamental to photography. It shapes how we perceive images by determining its realism, mood, depth, and overall visual quality. For anyone working with images, whether photographers, designers, or casual users, having control over illumination is essential for achieving the desired look. But lighting is hard to manipulate. It depends on scene geometry, surface materials, reflections, and shadows, all interacting in complex ways. Because illumination depends on complex physical interactions, explicit control typically requires access to geometric or environmental information that is rarely available in real-world images~\citep{basri2003lambertian}.

Recent diffusion models trained at scale have shown an ability to implicitly capture aspects of physical reasoning~\citep{wiedemer2025videomodelszeroshotlearners}, suggesting that illumination control may be learnable directly from data. Closed-sourced models such as ImageGen~\citep{team2023gemini} already demonstrate impressive text-guided control over lighting conditions~\citep{team2023gemini,bai2023qwen}. However, such models are proprietary which prevents either directly inspecting or extending them, or they are not transparent in terms of the dataset and methods used in training. Distillation from such models is also not optimal, as it often leads to limited generalization and constrains the student to a narrow synthetic distribution.

In contrast, existing open-source approaches to image relighting either rely on auxiliary control signals such as depth maps or normal maps~\citep{kocsis2024lightitilluminationmodelingcontrol}, or are not fully reproducible due to unreleased data and pipelines~\citep{iclight}. Other recent approaches require specialized conditioning inputs such as HDR environment maps~\citep{jin2024neural}, limiting accessibility for typical users. 

In this work, we frame illumination control as a supervised image editing problem conditioned on natural language instructions. Rather than collecting paired images of the same scene under different lighting, we construct supervision by synthetically degrading well-lit images. We introduce a fully open and reproducible data engine that transforms in-the-wild images into instruction-based relighting triplets, enabling diffusion models to relight subjects while preserving identity without requiring external control inputs.

To summarize, our main contributions are:
\begin{enumerate}
    \item We present a fully open and reproducible data engine that constructs instruction-based relighting supervision from in-the-wild images, combining CLIP-based filtering, intrinsic image decomposition, geometry-aware degradation, and automated light editing instruction generation using a vision--language model.
    \item We frame illumination control as a text-guided image editing problem by synthetically degrading well-lit images, eliminating the need for paired captures or auxiliary inputs such as depth maps.
    \item We show that fine-tuning a diffusion model on the resulting data achieves 2× better perceptual similarity and up to 17× better identity preservation compared to SD 1.5, SDXL, and FLUX.1-dev baselines, while generalizing to out-of-distribution images from CelebA-HQ.
\end{enumerate}

\begin{figure*}[t]
  \centering
  \includegraphics[width=\textwidth]{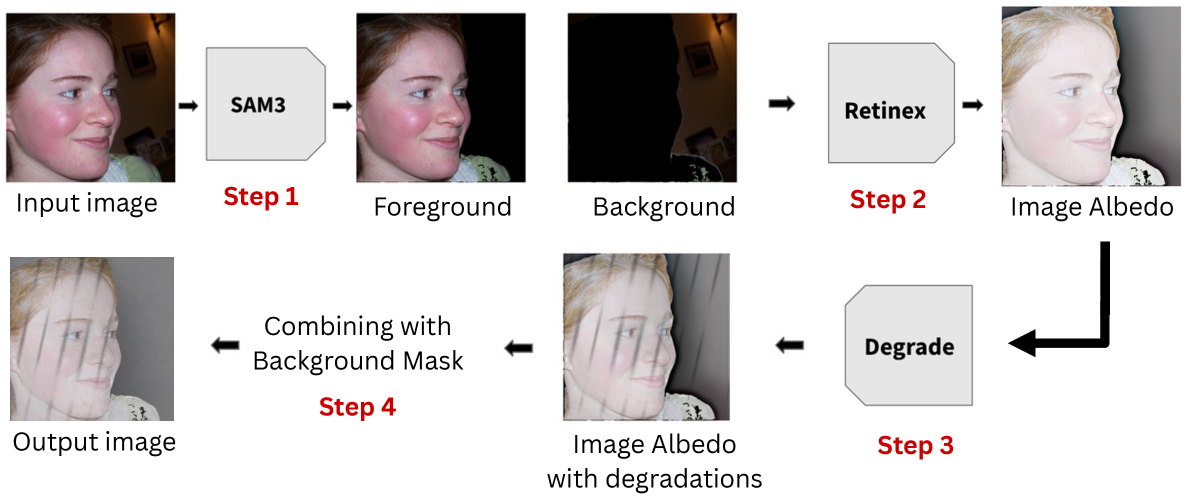}
  \caption{Overview of our data engine. Starting from a well-illuminated image, the pipeline filters for lighting quality, segments the subject, extracts albedo, applies synthetic degradation, and generates text instructions describing the target illumination}
  \label{fig:pipeline}
\end{figure*}

\section{Related Work}

\textbf{Image Relighting} Classical approaches to image relighting decompose the scene into geometry, materials, and lighting, then re-render under new illumination conditions~\citep{basri2003lambertian}. These methods require multi-view captures or specialized equipment like light stages~\citep{debevec2000acquiring}, and are limited to simplified reflectance models. Learning-based methods addressed some of these constraints by training on light stage data~\citep{sun2019single} or synthetic rendered images~\citep{zhou2019deep}. However, light stage capture is expensive and restricted to controlled settings. These approaches are typically limited to faces and rely on explicit decomposition into albedo, geometry, and shading, where errors in intermediate predictions accumulate~\citep{pandey2021total}.

\noindent\textbf{Diffusion Models For Image Editing} Diffusion models generate images by learning to reverse a gradual noising process~\citep{ho2020denoising}. Latent Diffusion Models made this practical by operating in a compressed latent space rather than pixel space, significantly reducing computational requirements~\citep{rombach2022high}. Stable Diffusion built on this approach, enabling large-scale text-to-image generation on consumer hardware~\citep{rombach2022high}. For image editing, InstructPix2Pix demonstrated that synthetic paired data can train models to follow natural language editing instructions without requiring per-example fine-tuning~\citep{brooks2023instructpix2pix}. ControlNet extended diffusion models with spatial control through conditioning signals like depth maps, edges, and pose, but requires users to provide these explicit control inputs~\citep{zhang2023adding}.

\noindent\textbf{Diffusion-Based Relighting} Recent work has explored using diffusion models to learn relighting end-to-end without explicit scene decomposition. IC-Light trains on large-scale diverse data and shows promising results, but does not release its pipeline or dataset~\citep{zhang2025scaling}. Neural Gaffer conditions on HDR environment maps, requiring specialized input that most users do not have~\citep{jin2024neural}. DreamLight supports both image and text-based relighting but introduces architectural complexity~\citep{liu2025dreamlight}. IC-Light is closest to our goal, but its lack of released code and data makes it non-reproducible. There remains no fully open-source, reproducible pipeline for text-guided illumination control.
\section{Methodology}

Given that high-quality images with good lighting are abundant on the web, our approach is to build a data engine that mines these images and processes them into supervised training triplets. Each triplet consists of a poorly-lit input image, a natural language instruction describing the target lighting, and a well-illuminated output image following that edit instruction.

The core challenge is that finding natural pairs of the same scene under different lighting conditions is impractical. However, well-illuminated images are readily available. Our solution is to work backwards: we start with well-lit images as our ground truth outputs, synthetically create poorly-lit versions as our inputs, and generate text descriptions of the original lighting as our edit instructions. 

\begin{figure*}[t]
  \centering
  \includegraphics[width=\textwidth]{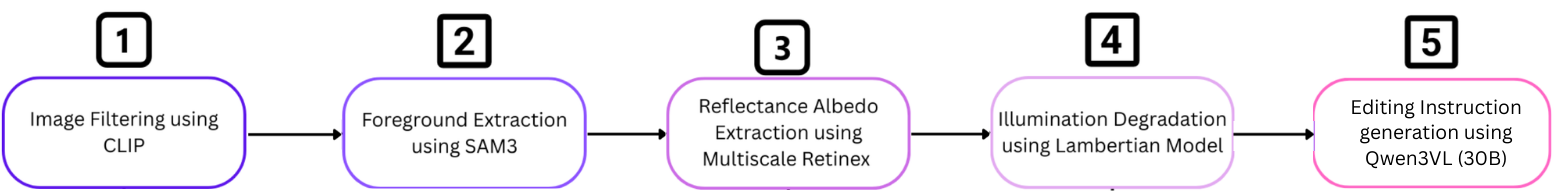}
  \caption{\textbf{Data engine pipeline.} Starting from a large image collection, we filter for well-illuminated images using CLIP, segment the subject with SAM~3, extract a lighting-neutral albedo via Multi-Scale Retinex, apply synthetic illumination degradation using depth-aware Lambertian shading, and then generate natural language lighting editing instructions with Qwen3-VL to complete each training triplet.}
  \label{fig:pipeline_purple}
\end{figure*}

As shown in Figure~\ref{fig:pipeline_purple}, our pipeline works as follows: we start with a large collection consisting of facial images and filter out those with poor illumination, keeping only well-lit images. Since our goal is to relight the subject, we segment them out from the background. We then remove traces of existing lighting from the subject to get a neutral starting point. Next, we apply synthetic shadows and lighting degradation to create the poorly-lit input. Finally, we generate natural language descriptions of the target lighting conditions. The following sub-sections describe each step in detail.

\begin{figure*}[t]
  \centering
  \includegraphics[width=\textwidth]{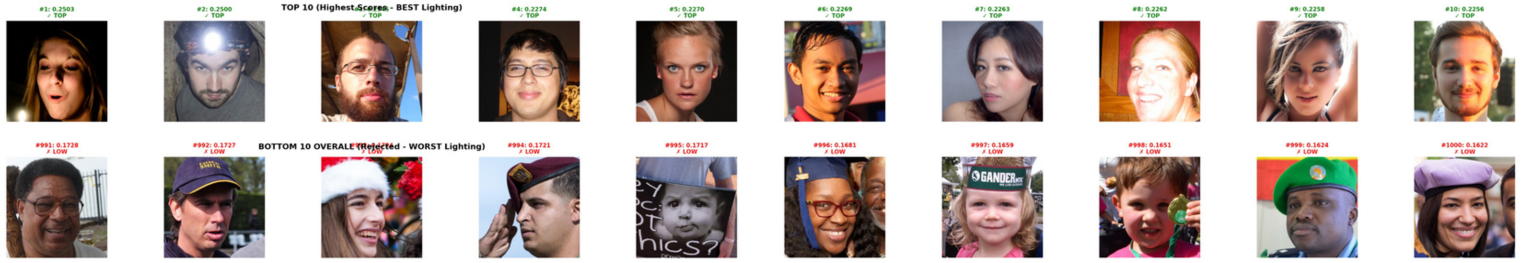}
  \caption{\textbf{CLIP-based illumination filtering.} Images scoring above our threshold of 0.21 (top row) exhibit clear, well-lit faces, while images below this threshold (bottom row) show poor illumination or occluded faces.}
  \label{fig:filtering}
\end{figure*}

\subsection{Filtering Illuminated Images}
\label{sec:filtering}

A key requirement of our data engine is a large collection of high-quality, well-lit images that can serve as reliable ground truth targets for relighting. We initially explored several publicly available facial image datasets, including large-scale web-scraped collections and celebrity datasets. However, many of these datasets exhibit significant variation in image resolution, compression artifacts, occlusions, or inconsistent and poor illumination, which complicates their use as ground truth for illumination control.

We ultimately select the Flickr-Faces-HQ (FFHQ) dataset~\citep{ffhq} as our primary image source. FFHQ contains 70,000 high-quality face images collected from Flickr, curated to exhibit substantial diversity in age, ethnicity, pose, and expression. Importantly for our task, FFHQ images are provided at a uniform resolution of $1024 \times 1024$ pixels and are predominantly well-lit, sharply focused, and minimally occluded. These properties make FFHQ particularly well-suited for learning fine-grained illumination effects and preserving subject identity during relighting.

Despite the overall quality of FFHQ, illumination conditions still vary across images. To filter for consistently well-lit images at scale, we utilize CLIP~\citep{radford2021learning} as a semantic scoring mechanism. Specifically, we use the CLIP ViT-B/32 model to compute image--text similarity scores between each image and a set of seven text prompts describing good lighting conditions, such as: ``beautiful lighting,'' ``professional lighting,'' ``well lit face,'' and ``bright and clear lighting.'' For each image, we compute the similarity score for each prompt and take their average to obtain a single scalar lighting quality score.

We then perform manual inspection across a range of similarity scores to determine a reliable threshold that separates well-lit images from poorly illuminated ones. We observe that images with an average CLIP similarity score greater than 0.21 consistently exhibit good lighting conditions and clear facial details. Based on this observation, we retain only images whose similarity score exceeds this threshold.

Applying this filtering procedure yields approximately 12,000 well-lit images from the FFHQ dataset. We split this subset into 10,000 training images, 1,000 validation images, and 1,000 test images. These filtered images serve as ground truth outputs in our instruction-based relighting triplets. Figure~\ref{fig:filtering} visualizes representative examples of images with the highest and lowest CLIP lighting similarity scores, illustrating how the proposed filtering criterion separates well-lit images from poorly illuminated ones.

\subsection{Foreground Segmentation}

Our goal is to manipulate illumination on the subject while preserving identity and intrinsic appearance. However, images collected in the wild often contain complex and diverse backgrounds with their own lighting cues, shadows, and color casts. If left unaddressed, these background lighting signals can interfere with both albedo extraction and subsequent synthetic relighting, leading to inconsistent or unrealistic results. Consequently, isolating the subject from the background is a critical step in our data engine.

To achieve this, we perform foreground segmentation to extract the subject region from each filtered image. We adopt SAM~3~\citep{sam3}, a recent iteration of Segment Anything Models~\citep{sam2}, which provides strong zero-shot generalization across a wide variety of scenes and supports natural language prompts. Unlike category-specific segmentation models, SAM~3 allows us to specify high-level semantic concepts such as ``person'' or ``face'' without requiring dataset-specific fine-tuning. For each image, we prompt SAM~3 with the text query \textit{``person''} and obtain a binary segmentation mask corresponding to the subject. This mask is used to isolate the foreground subject while suppressing background pixels. The resulting foreground image contains only the subject’s appearance, significantly reducing background-induced illumination artifacts in downstream processing.

Foreground segmentation serves two key purposes in our pipeline. First, it enables more accurate albedo extraction by preventing background lighting patterns from influencing the estimated reflectance of the subject. Second, it allows synthetic lighting and shadow effects to be applied exclusively to the subject during degradation. By decoupling subject lighting from background context, segmentation improves both the realism of synthetic degradations and the stability of supervision signals used for training. Figure~\ref{fig:pipeline} illustrates the role of foreground segmentation within the overall data engine pipeline, highlighting how subject isolation precedes albedo extraction and illumination degradation.

\subsection{Albedo Extraction}

Even after segmentation, the subject still contains significant illumination information in the form of shading, highlights, and cast shadows. Applying synthetic lighting directly on top of this existing illumination would result in compounding lighting effects and reduce realism. To obtain a neutral starting point for relighting, we estimate the subject's albedo: the intrinsic color and reflectance of the surface independent of illumination.

We adopt Retinex-based intrinsic image decomposition~\citep{land1977retinex} to separate the foreground image into reflectance and illumination components. Retinex theory models an image $I$ as the element-wise product of reflectance $R$ and illumination $L$, i.e.\ $I = R \odot L$, enabling the removal of low-frequency lighting variations while preserving high-frequency texture and color information. We apply this decomposition exclusively to the segmented foreground, ensuring that background pixels do not influence the estimated reflectance.

Specifically, we use Multi-Scale Retinex (MSR)~\citep{rahman1997multiscale}, which estimates reflectance by averaging log-domain differences between the image and its Gaussian-blurred versions across multiple spatial scales ($\sigma \in \{15, 80, 250\}$), followed by per-channel color normalization. To mitigate over-brightening that can result from aggressive illumination removal, we blend the estimated albedo with the original foreground image using a randomly sampled ratio $\alpha \in [0.15, 0.25]$. This blending preserves natural skin tones while still substantially reducing the influence of the original illumination, and introduces controlled variability across the dataset.

\begin{figure*}[t]
  \centering
  \includegraphics[width=\textwidth]{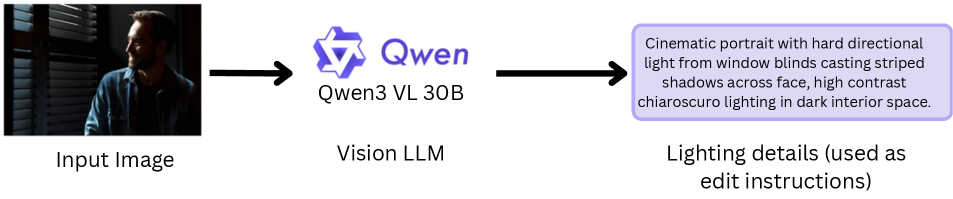}
  \caption{Editing instruction generation. We use Qwen3-VL to generate natural language descriptions of lighting conditions, which serve as text instructions for our training triplets.}
  \label{fig:instruction_generation}
\end{figure*}

\subsection{Illumination Degradation}

We now have a lighting-neutral albedo of the subject, which serves as our starting point for creating the degraded input image. The goal is to synthesize realistic and diverse poor lighting conditions that are physically plausible and respect the subject's geometry.

To achieve geometry-aware shading, we estimate a monocular depth map for each image using MiDaS~\citep{ranftl2020midas} and derive approximate surface normals from spatial depth gradients. Using a Lambertian shading model~\citep{basri2003lambertian}, we simulate directional lighting by sampling a random light direction over the upper hemisphere and computing shading as the dot product between the surface normal and the light direction.  We include an ambient illumination term to avoid overly harsh shadows, preserving global visibility while maintaining directional contrast.

In addition to geometry-based shading, we overlay procedural shadow patterns to simulate cast shadows from occluding objects. We implement ten pattern generators, including venetian blinds, window frames, tree foliage (via fractal Brownian motion), branches, curtains, fences, and architectural screens. For each image, a pattern is randomly selected according to a weighted distribution, Gaussian-blurred for realism, and composited onto the shaded image at a random opacity in $[0.35, 0.6]$. The subject is then placed on a neutral gray background to remove background lighting cues. By independently varying light direction, ambient intensity, pattern type, and overlay strength across images, we generate a broad distribution of degradation conditions that encourages the model to learn diverse illumination transformations.

\subsection{Instruction Generation}

The final component of each training triplet is a natural language instruction describing the target illumination. We automate this using Qwen3-VL~\citep{qwen3vl}, a multimodal vision--language model. For each well-lit ground truth image, we prompt the model to produce a single descriptive sentence capturing the lighting direction, quality, and atmosphere (the full prompt template is provided in Appendix~\ref{sec:vlm_prompt}). This yields diverse descriptions such as \textit{``soft natural daylight illuminating the face from the front-left''} or \textit{``dramatic side lighting casting deep shadows across half the face.''} Figure~\ref{fig:instruction_generation} illustrates the instruction generation stage, showing how target lighting descriptions are produced from well-lit ground truth images to complete each training triplet.

Importantly, instructions are generated solely from the ground truth image and are never conditioned on the degraded input, ensuring each instruction describes the desired target illumination rather than artifacts in the input. Together with the degraded input and well-lit output, these instructions complete the supervised training triplets. Algorithm~\ref{alg:data_engine} summarizes the full data engine pipeline used to construct the training triplets.

\section{Experimental Setup}
\begin{algorithm}[t]
\caption{Illumination Control Data Engine Pipeline}
\label{alg:data_engine}
\begin{algorithmic}[1]
\Require Large-scale Raw Image Dataset $\mathcal{D}_{raw}$
\Ensure Supervised Training Triplets $\mathcal{D}_{train} = \{(x_{deg}, t_{instr}, x_{gt})\}$

\State Initialize $\mathcal{D}_{train} \leftarrow \emptyset$

\ForAll{image $I \in \mathcal{D}_{raw}$}
    \State \text{\Comment 1. Filter: Check illumination quality (Sec.~3.1)}
    \State $s_{score} \leftarrow \text{CLIP}(I,\ \text{"lighting prompts"})$
    \If{$s_{score} > 0.21$}
        \State $x_{gt} \leftarrow I$ \Comment{Set as Ground Truth}

        \State \text{\Comment 2. Segmentation: Isolate subject (Sec.~3.2)}
        \State $M_{mask} \leftarrow \text{SAM3}(x_{gt},\ \text{"face/person"})$

        \State \text{\Comment 3. Albedo: Remove existing lighting (Sec.~3.3)}
        \State $A_{albedo} \leftarrow \text{Retinex}(x_{gt} \odot M_{mask})$

        \State \text{\Comment 4. Degradation: Apply synthetic lighting (Sec.~3.4)}
        \State $D_{depth} \leftarrow \text{EstimateDepth}(x_{gt})$
        \State $x_{deg} \leftarrow \text{Lambertian}(A_{albedo}, D_{depth}) + \text{Shadows}$

        \State \text{\Comment 5. Instruction: Describe target lighting (Sec.~3.5)}
        \State $t_{instr} \leftarrow \text{Qwen3-VL}(x_{gt})$

        \State \text{\Comment Save Triplet}
        \State $\mathcal{D}_{train} \leftarrow \mathcal{D}_{train} \cup \{(x_{deg}, t_{instr}, x_{gt})\}$
    \EndIf
\EndFor









\State \Return $\mathcal{D}_{train}$
\end{algorithmic}
\end{algorithm}


\subsection{Dataset Construction}
Starting from the FFHQ dataset~\citep{ffhq}, we apply CLIP-based illumination filtering as described in Section~\ref{sec:filtering}, retaining images with an average CLIP similarity score above 0.21. This results in approximately 12,000 well-lit face images, which we split into 10,000 training, 1,000 validation, and 1,000 test images. Each image is processed through the full data engine to produce a triplet consisting of a degraded input image, a natural language lighting instruction, and a well-lit ground truth output image. Although FFHQ provides images at $1024 \times 1024$, we resize all images to $512 \times 512$ for training and evaluation, consistent with standard Stable Diffusion practices. To evaluate generalization beyond the FFHQ distribution, we additionally curate a qualitative test set of 64 images from the CelebA-HQ dataset~\citep{karras2018progressive} paired with diverse editing instructions spanning a wide range of lighting scenarios. This out-of-distribution set is used for qualitative comparisons in addition to our 1000 image test set.

\subsection{Model and Training Configuration}
We adopt the InstructPix2Pix architecture~\citep{brooks2023instructpix2pix} built on Stable Diffusion 1.5~\citep{rombach2022high} as our base model. The model takes as input a degraded image and a textual instruction and outputs a relit image conditioned on the instruction. During training, we freeze the variational autoencoder (VAE) and the text encoder, and fine-tune only the U-Net backbone. We train the model for 250 epochs using the AdamW optimizer with a learning rate of $1 \times 10^{-5}$ and per-GPU batch size of 24. All experiments are conducted at a resolution of $512 \times 512$. Training completes in approximately 5.5 hours on 8$\times$NVIDIA A100 80GB GPUs.

\subsection{Evaluation Protocol}
We evaluate our method on the held-out test set of 1,000 images and compare against three pretrained baselines: Stable Diffusion 1.5 (SD~1.5), Stable Diffusion XL (SDXL)~\citep{podell2023sdxl}, and FLUX.1-dev~\citep{flux2024}. All baselines are evaluated using their respective image-to-image pipelines, receiving the same degraded input images and lighting instructions as our model.

To capture different aspects of relighting quality, we utilize four complementary metrics. LPIPS~\citep{zhang2018unreasonable} measures perceptual similarity between the generated and ground truth images using deep features, where lower is better. SSIM~\citep{wang2004image} evaluates structural similarity based on luminance, contrast, and spatial structure, where higher is better. CLIP Score~\citep{hessel2021clipscore} measures text--image alignment between the generated output and the lighting instruction. Identity Score~\citep{deng2019arcface} evaluates identity preservation by computing cosine similarity between ArcFace embeddings of the generated image and the ground truth. Together, these metrics provide a comprehensive evaluation of perceptual quality, structural fidelity, instruction adherence, and identity preservation for the relighting task.

\section{Results}
We compare against three pretrained baselines without relighting-specific supervision: Stable Diffusion 1.5 (SD~1.5), Stable Diffusion XL (SDXL), and FLUX.1-dev, all using the same degraded inputs and lighting instructions.

\begin{figure*}[t]
  \centering
  \includegraphics[width=\textwidth]{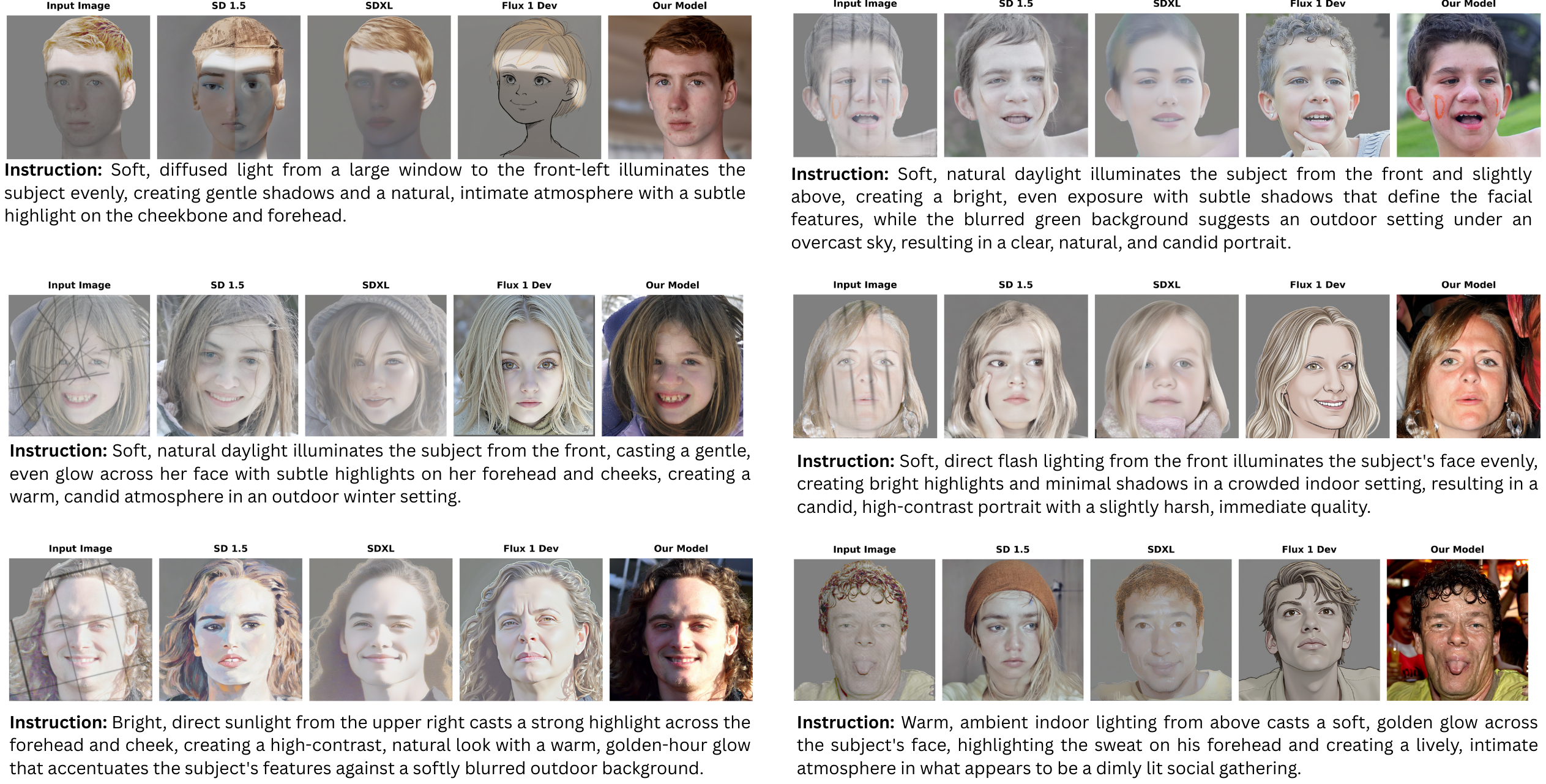}
  \caption{\textbf{Qualitative comparison on our FFHQ test set.} Given degraded inputs and lighting instructions, our model produces realistic relighting while preserving subject identity. All three baselines largely disregard the editing instruction and fail to maintain facial identity.}
  \label{fig:qualitative_results}
\end{figure*}

\subsection{Quantitative Results}
\begin{table}[t]
    \centering
    \setlength{\tabcolsep}{8pt}
    \renewcommand{\arraystretch}{1.25}
    \small
    \begin{tabular}{lllll}
        \toprule
        \textbf{Metric} & \textbf{SD 1.5} & \textbf{SDXL} & \textbf{FLUX.1-dev} & \textbf{Our Model} \\
        \midrule
        LPIPS $\downarrow$        & $0.6346 \pm 0.0901$   & $0.6292 \pm 0.0896$   & $0.6504 \pm 0.0787$   & $\mathbf{0.3002 \pm 0.0904}$   \\
        SSIM $\uparrow$           & $0.3802 \pm 0.0951$   & $0.4333 \pm 0.1009$   & $0.3726 \pm 0.0974$   & $\mathbf{0.5667 \pm 0.1002}$   \\
        CLIP $\uparrow$           & $\mathbf{0.2601 \pm 0.0280}$ & $0.2567 \pm 0.0291$ & $0.2520 \pm 0.0303$ & $0.2504 \pm 0.0314$           \\
        Identity Score $\uparrow$ & $0.0712 \pm 0.0788$   & $0.1088 \pm 0.0980$   & $0.0437 \pm 0.0796$   & $\mathbf{0.7591 \pm 0.1823}$   \\
        \bottomrule
    \end{tabular}
    \caption{\textbf{Quantitative Results.} Comparison of our model against SD\,1.5, SDXL, and FLUX.1-dev baselines across
    perceptual (LPIPS), structural (SSIM), text-alignment (CLIP),
    and identity preservation metrics. \textbf{Bold} indicates best
    performance per metric.}
    \label{tab:quantitative_results}
\end{table}

Table~\ref{tab:quantitative_results} reports quantitative comparisons across four complementary metrics on the 1,000-image held-out test set.

Our method substantially outperforms all three baselines on three of four metrics. In perceptual similarity, we achieve an LPIPS of 0.30, compared to 0.63 for SD~1.5, 0.63 for SDXL, and 0.65 for FLUX.1-dev. Structural consistency follows a similar trend: our SSIM of 0.57 outperforms SD~1.5 (0.38), SDXL (0.43), and FLUX.1-dev (0.37). Notably, despite being a significantly larger and more recent model, FLUX.1-dev performs comparably to or worse than the SD~1.5 baseline on all metrics, suggesting that model scale alone does not address the relighting task without task-specific supervision.

Identity preservation exhibits the largest gap. Our model achieves an Identity Score of 0.76, while SD~1.5 scores 0.07, SDXL scores 0.11, and FLUX.1-dev scores just 0.04. This $7{-}17{\times}$ improvement indicates that generic diffusion models frequently alter facial identity when following lighting instructions, whereas our relighting-specific supervision effectively disentangles illumination changes from identity-preserving appearance factors.

In CLIP Score, the baselines perform marginally higher (0.26 for SD~1.5 vs.\ 0.25 for ours). We attribute this to a fundamental trade-off: the baselines aggressively regenerate the image to match the text prompt, achieving slightly higher text--image alignment but at the cost of losing the input identity and structure. Our model instead learns to \emph{modify} the existing image's lighting, which inherently constrains the output to remain close to the input. The near-parity in CLIP Score confirms that our model follows lighting instructions comparably, while the large gains in LPIPS, SSIM, and Identity Score demonstrate that it does so while preserving the subject.

\begin{figure*}[t]
  \centering
  \includegraphics[width=0.95\textwidth]{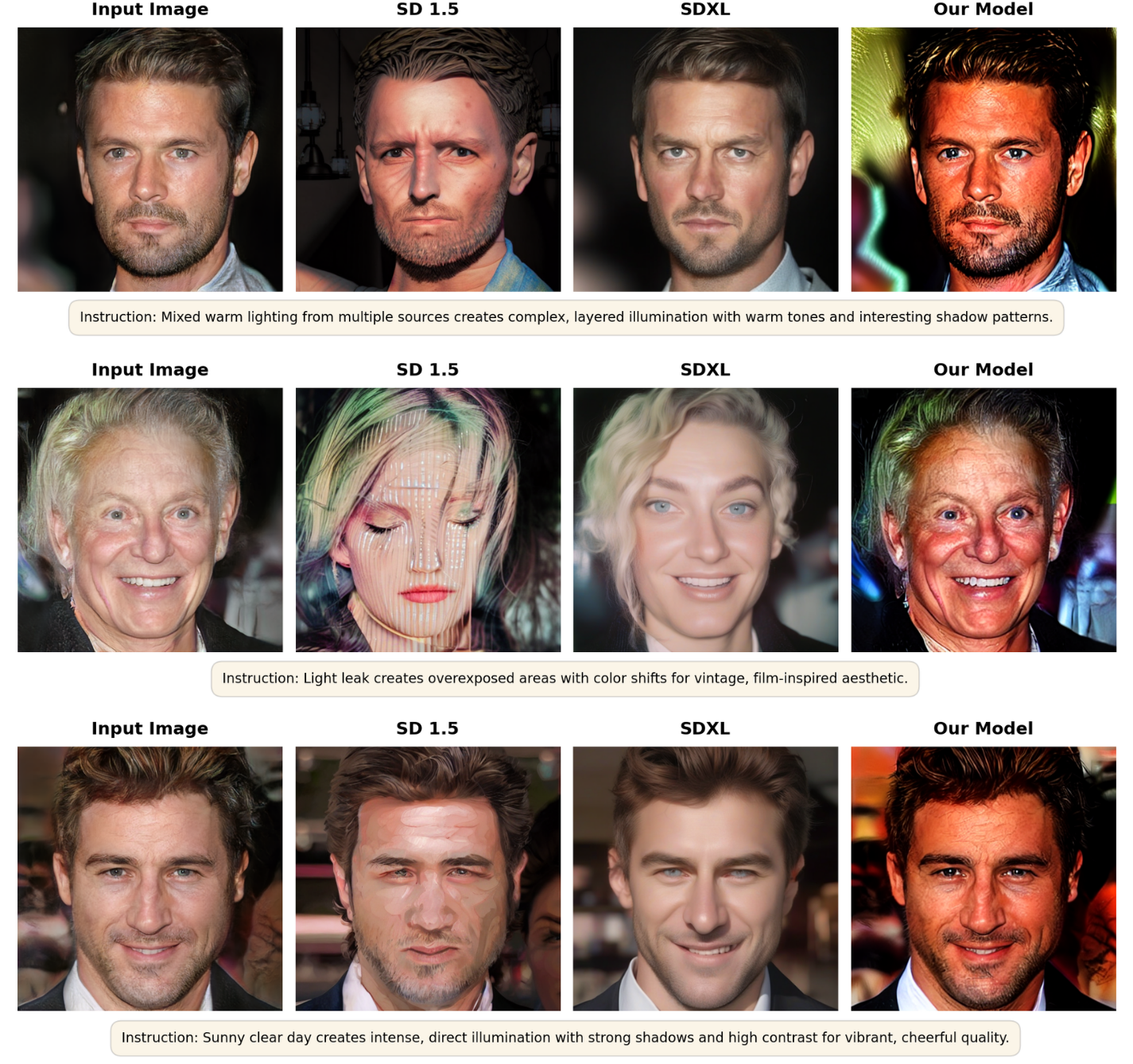}
  \caption{\textbf{Out-of-distribution generalization on CelebA-HQ.} Our model generalizes to unseen faces with diverse lighting instructions, while baselines exhibit inconsistent illumination and fail to preserve facial identity.}
  \label{fig:qual_celeb}
\end{figure*}

\subsection{Qualitative Results}
Figure~\ref{fig:qualitative_results} presents qualitative comparisons on the held-out FFHQ test set between our method and all three baselines under diverse lighting instructions. Each example shows the degraded input, the outputs of SD 1.5, SDXL, and FLUX.1-dev, and the output of our model. Across all examples, the baselines frequently introduce identity drift, distorted facial features, and inconsistent lighting effects. In contrast, our method produces illumination changes that closely match the specified lighting instruction while maintaining facial identity and fine-grained details.

Figure~\ref{fig:qual_celeb} shows qualitative results on the out-of-distribution CelebA-HQ, where the model encounters unseen faces with diverse lighting instructions. Our model generalizes effectively, producing plausible relighting while preserving identity. The baselines continue to exhibit significant identity drift and inconsistent lighting on this set as well.

These qualitative results complement the quantitative findings, illustrating that our data engine enables diffusion models to learn illumination-specific transformations rather than generic image regeneration. Together, the results demonstrate that instruction-based relighting supervision derived from synthetic degradation provides an effective and scalable solution for controllable illumination editing.

\section{Conclusion, Limitations, and Future Work}

We presented a fully open-source pipeline for learning illumination control in diffusion models by reframing relighting as an instruction-based image editing problem. Our data engine transforms well-lit in-the-wild images into supervised training triplets through CLIP-based filtering, foreground segmentation, Retinex-based albedo extraction, geometry-aware Lambertian degradation, and automated instruction generation via a vision--language model. 
Fine-tuning an SD 1.5 model on 10,000 such triplets yields $2{\times}$ better perceptual similarity (LPIPS: 0.30 vs.\ 0.63–0.65) and up to $17{\times}$ better identity preservation (Identity Score: 0.76 vs.\ 0.04–0.11) compared to SD~1.5, SDXL, and FLUX.1-dev baselines, while generalizing to out-of-distribution faces from CelebA-HQ. These results demonstrate that carefully designed synthetic supervision can teach diffusion models illumination-specific transformations rather than generic image regeneration. 

Our current pipeline focuses on human faces, leveraging the consistency of the FFHQ dataset, which limits direct applicability to other object categories and full-scene relighting. Extending the data engine to general scenes and exploring more expressive diffusion backbones are promising directions for future work. Additionally, supporting spatially localized or multi-light instructions could enable finer-grained relighting control. We have released our complete data engine, code and model weights to support reproducible research in this area.

\bibliography{iclr2026_conference}
\bibliographystyle{iclr2026_conference}

\appendix

\section{Appendix}

\subsection{Baseline Models}
\label{sec:baselines}

We compare our method against three pretrained diffusion models, all evaluated using their respective image-to-image pipelines to ensure a fair comparison. Each baseline receives the same degraded input image and natural language lighting instruction as our model.

\textbf{Stable Diffusion 1.5 (SD~1.5)}~\citep{rombach2022high} is a latent diffusion model with a U-Net backbone operating in the latent space of a pretrained variational autoencoder. It uses a single CLIP ViT-L/14 text encoder for conditioning and was trained on the LAION-5B dataset. We use the \texttt{StableDiffusionImg2ImgPipeline} from the Diffusers library, which takes a source image and a text prompt and generates an edited image by denoising from a partially noised version of the input.

\textbf{Stable Diffusion XL (SDXL)}~\citep{podell2023sdxl} is a significantly larger latent diffusion model that employs a larger U-Net architecture and dual text encoders (CLIP ViT-L/14 and OpenCLIP ViT-bigG/14) for improved text understanding and image fidelity. We use the \texttt{StableDiffusionXLImg2ImgPipeline}, which follows the same image-to-image paradigm as SD~1.5.

\textbf{FLUX.1-dev}~\citep{flux2024} is a recent open-weight model from Black Forest Labs based on flow matching rather than the traditional DDPM denoising framework. It represents the current state of the art in open-source text-to-image generation. We evaluate it using the \texttt{FluxImg2ImgPipeline} to maintain consistency with the image-to-image evaluation setup.

\textbf{InstructPix2Pix}~\citep{brooks2023instructpix2pix} is the architecture we adopt for fine-tuning. It extends Stable Diffusion 1.5 by concatenating the input image latent with the noisy latent as additional conditioning channels, enabling the model to follow natural language editing instructions while preserving the input image structure. We fine-tune this architecture on our relighting triplets as described in Section~4.2.

\subsection{Dataset Details}
\label{sec:datasets}

\textbf{FFHQ.} The Flickr-Faces-HQ (FFHQ) dataset~\citep{ffhq} contains 70,000 high-quality face images crawled from Flickr. Images are aligned and cropped to $1024 \times 1024$ pixels and exhibit substantial diversity in age, ethnicity, pose, expression, accessories, and background. The dataset was collected with a focus on quality and variation, making it well-suited for learning fine-grained facial appearance transformations. We use FFHQ as the source for our data engine pipeline, applying CLIP-based filtering (Section~3.1) to obtain approximately 12,000 well-lit images, which are split into 10,000 training, 1,000 validation, and 1,000 test images. All images are resized to $512 \times 512$ for training and evaluation.

\textbf{CelebA-HQ.} The CelebA-HQ dataset~\citep{karras2018progressive} contains 30,000 high-resolution celebrity face images derived from the original CelebA dataset through a multi-step quality enhancement pipeline. Images are provided at $1024 \times 1024$ resolution. CelebA-HQ differs from FFHQ in its distribution: it is derived from celebrity photographs with different capture conditions, backgrounds, and demographic composition. We curate a subset of 64 images from CelebA-HQ and pair them with diverse lighting instructions to serve as an out-of-distribution qualitative test set for evaluating generalization (Section~4.1).

\subsection{Lighting Description Generation Prompt}
\label{sec:vlm_prompt}

\begin{figure*}[t]
  \centering
  \includegraphics[width=\textwidth]{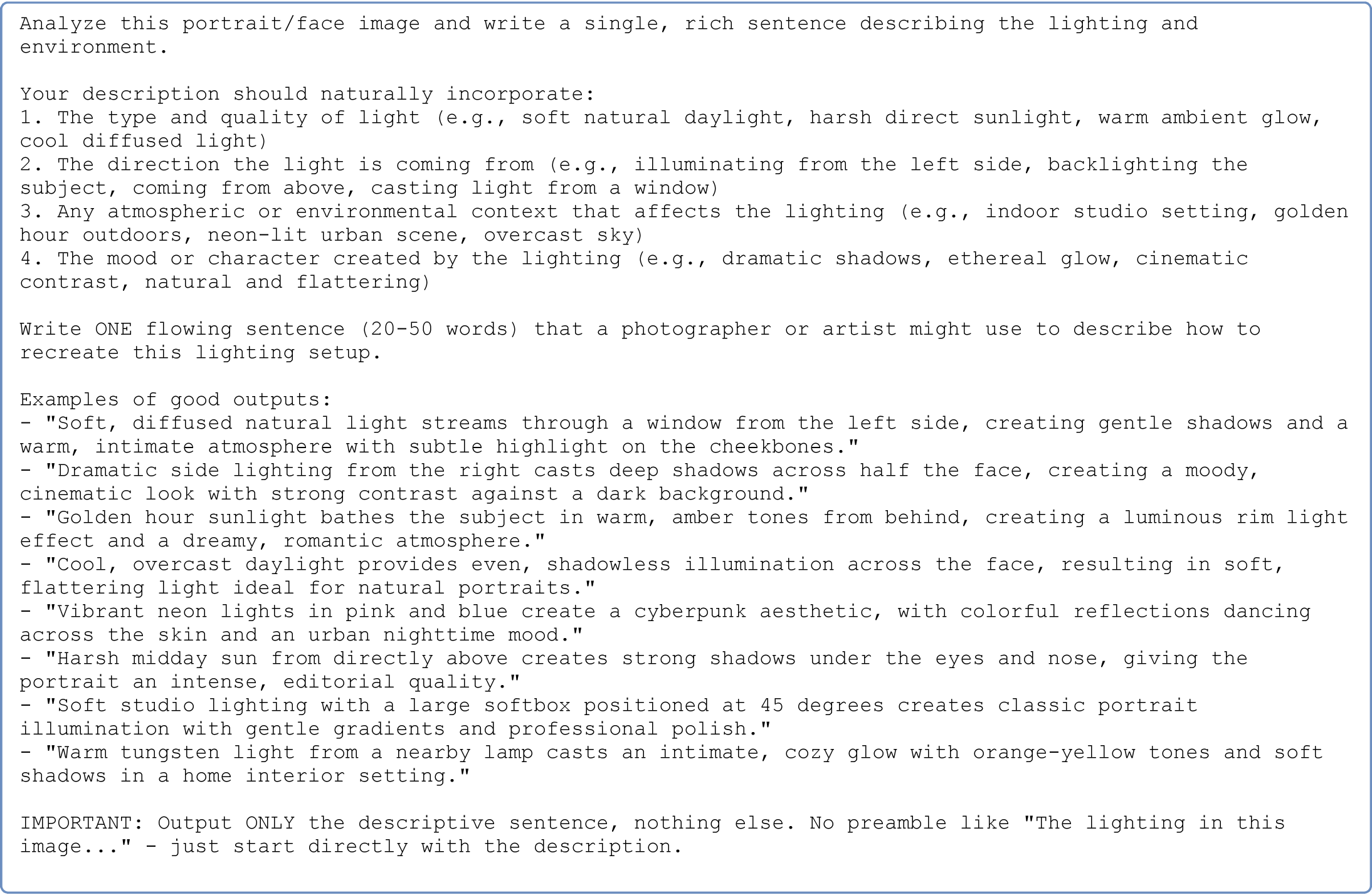}
  \caption{Prompt used for generating lighting descriptions.}
  \label{fig:prompt}
\end{figure*}

The following prompt in Figure~\ref{fig:prompt} is used with Qwen3-VL~\citep{qwen3vl} to generate lighting instructions for each ground truth image. We include structured guidance and few-shot examples to elicit diverse, photographer-style descriptions of illumination conditions.

\end{document}